\documentclass[conference]{IEEEtran}
\IEEEoverridecommandlockouts
\usepackage{cite}
\usepackage{amsmath,amssymb,amsfonts}
\usepackage[colorlinks,linkcolor=red]{hyperref}
\usepackage{algorithmic}
\usepackage{graphicx}
\graphicspath{{./figures/}}
\usepackage{textcomp}
\usepackage{xcolor}
\usepackage{balance}
\usepackage{geometry}
\usepackage{multicol} 
\usepackage{fancyhdr}
\usepackage{diagbox}
\usepackage{makecell}

\def\BibTeX{{\rm B\kern-.05em{\sc i\kern-.025em b}\kern-.08em
		T\kern-.1667em\lower.7ex\hbox{E}\kern-.125emX}}
\begin{document}
	
	\title{Feature Chirality in Deep Learning Models}

	\makeatletter
	\newcommand{\linebreakand}{%
	\end{@IEEEauthorhalign}
	\hfill\mbox{}\par
	\mbox{}\hfill\begin{@IEEEauthorhalign}
	}
	\makeatother
	
	\author{\IEEEauthorblockN{Shipeng Ji}
		\IEEEauthorblockA{\textit{\small{Command \& Control Engineering College}} \\
			\textit{Army Engineering University of PLA}\\
			Nanjing, China \\
			jishipeng19@foxmail.com}
		\and
		\IEEEauthorblockN{Yang Li}
		\IEEEauthorblockA{\textit{\small{Command \& Control Engineering College}}\\
			\textit{Army Engineering University of PLA}\\
			Nanjing, China \\
			solarleeon@outlook.com}
		\and
		\IEEEauthorblockN{Ruizhi Fu}
		\IEEEauthorblockA{
			\textit{No. 32316 Units of PLA}\\
			Urumqi, China\\
			xjfrz2020@163.com}
		\linebreakand
		\IEEEauthorblockN{Jiabao Wang}
		\IEEEauthorblockA{\textit{\small{Command \& Control Engineering College}} \\
			\textit{Army Engineering University of PLA}\\
			Nanjing, China \\
			jiabao\_1108@163.com}
		\and
		\IEEEauthorblockN{Zhuang Miao* \thanks{*Corresponding author.}}
		\IEEEauthorblockA{\textit{\small{Command \& Control Engineering College}} \\
			\textit{Army Engineering University of PLA}\\
			Nanjing, China \\
			emiao\_beyond@163.com}
		\and
	}
	
	\maketitle
	\thispagestyle{fancy}
	\fancyhead{}
	\lhead{}

	\cfoot{}
	\rfoot{}
	
	\begin{abstract}
		As deep learning applications extensively increase by leaps and bounds, their interpretability has become increasingly prominent. As a universal property, chirality exists widely in nature, and applying it to the explanatory research of deep learning may be helpful to some extent. Inspired by a recent study that used CNN (convolutional neural network), which applied visual chirality, to distinguish whether an image is flipped or not. In this paper, we study feature chirality innovatively, which shows how the statistics of deep learning models' feature data are changed by training. We rethink the feature-level chirality property, propose the feature chirality, and give the measure. Our analysis of feature chirality on AlexNet, VGG, and ResNet reveals similar but surprising results, including the prevalence of feature chirality in these models, the initialization methods of the models do not affect feature chirality. Our work shows that feature chirality implies model evaluation, interpretability of the model, and model parameters optimization.
		
	\end{abstract}
	
	
	\section{Introduction}
	During the last decade, deep learning methods have changed our daily lives from many perspectives \cite{alzubaidi2021review}. As a representative model in the field of computer vision, deep learning models achieve start-of-the-art performance on many tasks, such as object detection \cite{zou2019object} and semantic segmentation \cite{hao2020brief}. Despite such success, there is still one weakness of the deep learning model, which is the lack of explainability \cite{zhang2018interpretable}.
	
	Recently, many researchers try to explain deep learning models from a different perspective, such as saliency maps \cite{boyd2022human} and disentangled representations \cite{locatello2019challenging}. As a new way of disentangled representations, visual chirality \cite{lin2020visual} was first proposed to distinguish whether an image is flipped or not. Then, visual chirality was also used to freehand sketch recognition \cite{zheng2021visual}, mirror detection \cite{tan2022mirror}, and expression recognition \cite{lo2022facial}.
	
	It is well known that chirality is a universal law of nature, which appears widely in the winding direction of plants in biology \cite{darwin1875movements}, the spin direction of atmospheric cyclones in meteorology \cite{kumar2021intriguing}, chiral compounds in chemistry \cite{ni2021chirality}, parity non-conservation in physics \cite{lee1956question, wu1957experimental}, etc. Visual chirality provides the first validation of how this universal law relates to the image distribution in a dataset. Visual chirality is the statistical change in image distribution brought about by image flipping in data augmentation. Inspired by this work, we set out to consider whether chirality could be applied to the parameter analysis of deep learning models to perform discriminant analysis of unknown models, and then obtain special properties between models.
	
	\begin{figure}[hbtp]
		\centerline{\includegraphics[width=3in]{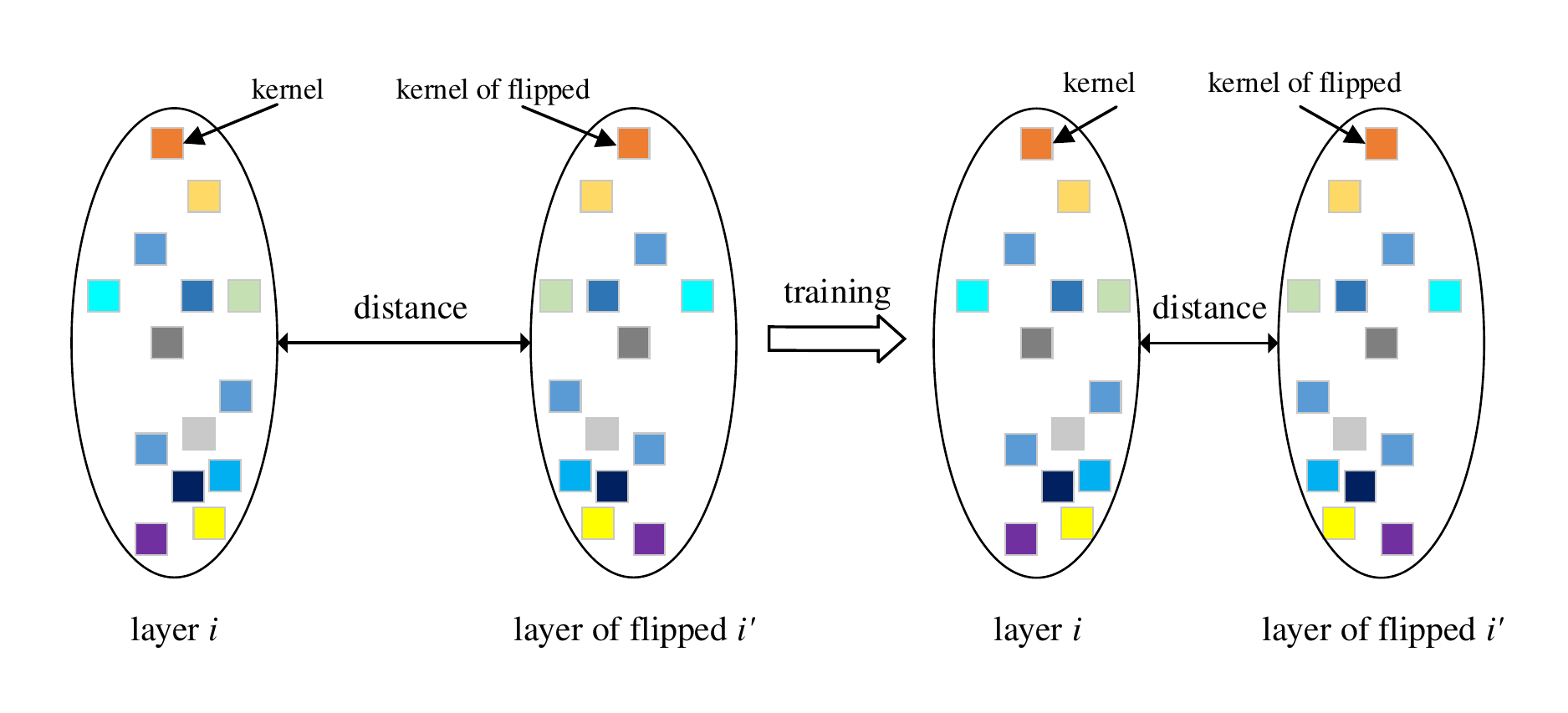}}
		\caption{The interpretation of feature chirality. For a model with feature chirality, the distance between the convolution layer and its flipped layer will decrease after training.}
		\label{fc}
	\end{figure}
	
	To achieve the above goal, the concept of feature chirality is first introduced in this paper, to represent the variation of similarity in model parameters from the chiral perspective. This variation directly reflects the impact of the training process on the parameters of the deep learning model and implies the changes in the deep learning model after learning useful knowledge in the training dataset. To our best knowledge, this is the first study to analyze the model parameters from the feature chirality perspective. After training, the internal similarity of the parameters in the model changes significantly, which is referred to as feature chirality. As shown in Fig. \ref{fc}, for a model with feature chirality, the similarity of model parameters changes significantly after training. Then, we design a measure of feature chirality to analyze the chiral perspective of deep learning model parameters by computing the average kernel similarity in the convolutional layer. Our analysis of ten classical models yielded some unexpected but reasonable conclusions, including the feature chirality of the deep learning models is surprisingly obvious; the initialization method does not affect the initial kernel similarity of the model; the results of the same series network are consistent, indicating the convolution kernel similarity of the same series network has a similar "fingerprint". These conclusions are interesting on topics ranging from the interpretability of deep learning to model evaluation. For instance, for the deep learning model "fingerprint" we found, we can easily identify which series a model belongs to in the black box state, and determine whether it has been trained without running the model.
	
	The main contributions of this paper can be summarized as follows:
	
	\begin{itemize}
		\item We observe the feature chirality phenomenon on deep learning models’ parameters and design an intuitive measure of it to observe its existence.
		
		\item This is the first study to explore the changes of kernel parameters in deep learning models before and after training from a chiral perspective.
		
		\item Extensive experimental analysis results on AlexNet \cite{krizhevsky2017imagenet}, VGG \cite{simonyan2014very}, and ResNet \cite{he2016deep} show that the feature chirality is ubiquitous in deep learning models, and it is found that the initialization methods do not affect the initial kernel similarity of the model and the variation trend of kernel similarity of the same series network is consistent.
		
	\end{itemize}
	
	\section{Related Work}\label{second}
	
	Chirality is one of the most important and widespread phenomena in nature. In the early years, naturalists and biologists such as Darwin and Wallace paid great attention to the macroscopic chirality of biological beings. Darwin \cite{darwin1875movements} described the chiral phenomenon of 42 kinds of climbing plants in the \textit{Movement and Habits of Climbing Plants}. However, enantiomers were not found until the discovery of polarized light by Louis Pasteur \cite{freemantle2003chemistry} studying the optical rotation of ammonium sodium tartrate crystals and their aqueous solutions. Later discovery that the arrangement of the corresponding atoms in space between left-handed and right-handed enantiomers is a mirror image of each other, just as a person's left and right hands are mirror images of each other. That is where the idea of the chiral and chiral molecule was introduced. Kelvin \cite{kelvin1894molecular} defined chirality as: “any geometrical figure, or group of points, its image in a plane mirror, ideally realized, cannot be brought to coincide with itself.” That is, a chiral object and its mirror image cannot be twisted or rotated in any way to coincide.
	
	Symmetry is one of the enduring topics in computer vision. To some extent, it can be considered as chirality, because an object with symmetry does not have chirality. Pickup \cite{pickup2014seeing} and Wei \cite{wei2018learning} study the asymmetry of time in videos - time chirality (i.e., the arrow of time), to find out what makes a video appears to be playing forward or backward. Recently, Lin \cite{lin2020visual} studied spatial asymmetry of chirality in images - spatial chirality (i.e., visual chirality), looking for what makes images seem normal or mirrored.
	
	As an important way of data augmentation, flipping is widely used in deep learning. However, mirror images represent an object (e.g., inverted text) that does not exist in the real world. To answer the question of whether and how such samples affect image distribution on the scale, Lin \cite{lin2020visual} proposed a self-supervised learning method to predict whether images are flipped and analyze the features of chiral cues. As the definition of the visual chirality, Lin introduced that the visual chirality exists if the \textit{commutative residual} $E\left( X \right)$ of any image $X$ in the image dataset is non-zero. $E\left( X \right)$ as follows:

	\begin{equation}
		E\left( X \right) = \left| {{\bf{J}}\left( {{\bf{T}}\left( X \right)} \right) - {\bf{T}}\left( {{\bf{J}}\left( X \right)} \right)} \right|
		\label{ejttj}
	\end{equation}
	
	\noindent where ${\bf{T}}$ is a flipping operation and ${\bf{J}}$ is an image processing operation (such as demosaicing and JPEG compression).
	
	Then he found some visual chiral cues in image distribution, such as text cues, high-level cues (e.g., mobile phones, guitars, and watches), and facial chiral cues (e.g., bangs on the forehead and beard). 
	
	Later, some researchers applied it to recognition and detection. Zheng \cite{zheng2021visual} directly applied visual chirality to freehand sketch recognition, Tan \cite{tan2022mirror} applied visual chiral cues to mirror detection, and Lo \cite{lo2022facial} proposed facial chirality and used visual chiral cues for expression recognition. These show the value of exploring chiral phenomena in the field of deep learning.
	
	The essential problem in our study is the interpretability of deep learning models. For feature analysis of deep learning models, some scholars analyze deep learning models' features globally \cite{yosinski2014transferable, aubry2015understanding, dong2017towards}, while others learn adversarial samples from deep learning models locally \cite{ross2018improving, su2019one, koh2017understanding}. However, they lack some of the aspects of chirality. In this paper, we analyze the features of the model using feature chirality. Because of that, we consider our work to be unique and novel while we inherit the latest chirality research in the field of computer vision from others.

	\section{Feature Chirality and Measure }\label{third}
	
	Visual chirality is introduced by Lin \cite{lin2020visual} to explain the asymmetry between the original image and its flipped image. Accordingly, this section defines feature chirality and its measure.
	
	Tan \cite{tan2019efficientnet} treats the convolutional layer as an operation that one way transforms the input into the output. Since the input, output, and operation in each convolution layer are different, the convolution layer $i$ can be described as follows:
	
	\begin{equation}
		{Y_i} = {F_i}\left( {{X_i}} \right)
		\label{yfx}
	\end{equation}

	\noindent where ${X_i}$, ${Y_i}$, and ${F_i}$ are the input tensor, the output tensor, and the convolution operation. This paper studies the changes of convolutional layer $i$ parameters ${L_i} = \left\langle {{B_i},{C_i},{H_i},{W_i}} \right\rangle$ after training under the influence of operation ${F_i}$, where ${B_i}$ is the number of kernels, ${C_i}$ is the channel dimension, ${H_i}$ and ${W_i}$ are the spatial dimension. 
	
	After training, the internal similarity of the parameters in the model changes significantly, which is referred to as feature chirality: \textit{the model has feature chirality if the average kernel similarity of the pre-trained model is significantly different from that of the untrained model; otherwise, the model has no feature chirality.} For the model $M$, ${L_i}$ is the parameters of convolutional layer $i$ of $M$. The model $M$ has feature chirality if the \textit{commutative residual} $E\left( M \right)$ of the deep learning model is non-zero (i.e., it satisfies Eq. (\ref{ee})). Fig. \ref{fc} is an example of a model with feature chirality.
	
	\begin{equation}
		{
			E\left( M \right) = \sum\limits_{i = 1}^N {E\left( {{L_i}} \right)}  \ne 0
		}
		\label{ee}
	\end{equation}
	
	\noindent where $N$ is the number of convolutional layers of the model $M$. 
	
	\begin{table*}[hbtp]
		\renewcommand{\arraystretch}{1.3}
		\caption{Architectures of AlexNet and VGG}
		\centering
		\begin{tabular}{cccccc}
			\hline
			Stage & AlexNet & VGG-11 & VGG-13 & VGG-16 & VGG-19 \\ \hline
			1     & $\left[ {11 \times 11,{\kern 1pt} 64} \right] \times 1$ & $\left[ {3 \times 3,64} \right] \times 1$  & $\left[ {3 \times 3,64} \right] \times 2$ & $\left[ {3 \times 3,64} \right] \times 2$ & $\left[ {3 \times 3,64} \right] \times 2$\\
			2     &  $\left[ {5 \times 5,192} \right] \times 1$ & $\left[ {3 \times 3,128} \right] \times 1$ & $\left[ {3 \times 3,128} \right] \times 2$ & $\left[ {3 \times 3,128} \right] \times 2$ &  $\left[ {3 \times 3,128} \right] \times 2$ \\
			3     & $\left[ {3 \times 3,384} \right] \times 1$ & $\left[ {3 \times 3,256} \right] \times 2$ & $\left[ {3 \times 3,256} \right] \times 2$ & $\left[ {3 \times 3,256} \right] \times 3$ &  $\left[ {3 \times 3,256} \right] \times 4$ \\
			4     & $\left[ {3 \times 3,256} \right] \times 1$ & $\left[ {3 \times 3,512} \right] \times 2$ & $\left[ {3 \times 3,512} \right] \times 2$ & $\left[ {3 \times 3,512} \right] \times 3$ &  $\left[ {3 \times 3,512} \right] \times 4$\\
			5     & $\left[ {3 \times 3,256} \right] \times 1$ & $\left[ {3 \times 3,512} \right] \times 2$ & $\left[ {3 \times 3,512} \right] \times 2$ & $\left[ {3 \times 3,512} \right] \times 3$ &  $\left[ {3 \times 3,512} \right] \times 4$ \\ \hline
		\end{tabular}
		\label{AAA}
	\end{table*}
	
	Inspired by the definition of visual chirality, we define feature chirality. $X$ in Eq. (\ref{ejttj}) is a two-dimensional image, while this paper studies the four-dimensional parameters of convolution layers, so Eq. (\ref{ejttj}) needs to be extended. The formal expression of the \textit{commutative residual} $E\left( {{L_i}} \right)$ is expressed as follows:
	
	\begin{equation}
		{E\left( {{L_i}} \right) = \left| {{\bf{S}}\left( {{L_i}} \right) - {\bf{S}}\left( {{{\bf{T}}_r}\left( {{L_i}} \right)} \right)} \right|}
		\label{eslstl}
	\end{equation}
	
	\noindent Since feature chirality is to compare the difference between pre-trained and untrained models, it is inconvenient to quantify them directly. Thus, we use the average kernel similarity of the model to compare the changes in the similarity of parameters within the convolution layers. We redefined ${\bf{J}}\left( {{\bf{T}}\left( X \right)} \right)$ and ${\bf{T}}\left( {{\bf{J}}\left( X \right)} \right)$ in Eq. (\ref{ejttj}) as ${\bf{S}}\left( {{L_i}} \right)$ and ${\bf{S}}\left( {{{\bf{T}}_r}\left( {{L_i}} \right)} \right)$ in Eq. (\ref{eslstl}), where ${\bf{S}}\left( {{L_i}} \right)$ is denoted as the operation of computing the average kernel similarity of the convolution layer $i$ in the untrained model, ${{\bf{T}}_r}\left( {{L_i}} \right)$ is denoted as the operation of training the model, and ${\bf{S}}\left( {{{\bf{T}}_r}\left( {{L_i}} \right)} \right)$ is denoted as the operation of computing the average kernel similarity of the convolution layer $i$ in the pre-trained model.
	
	\begin{figure}[hbtp]
		\centerline{\includegraphics[width=2in]{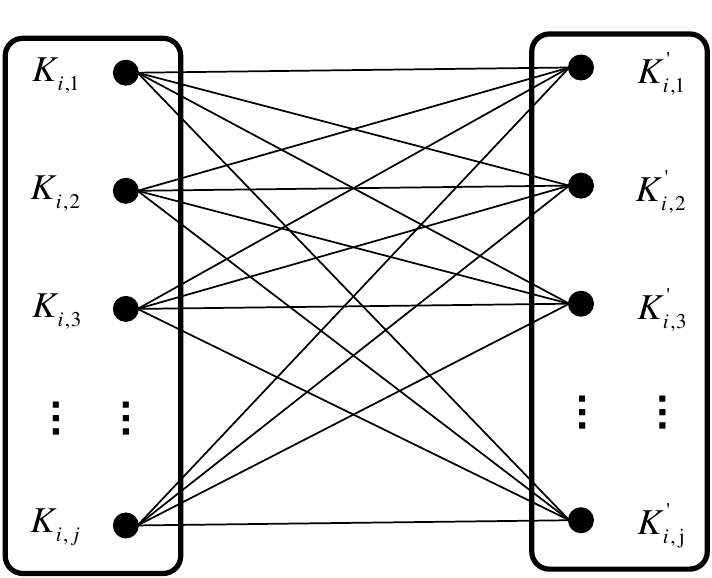}}
		\caption{Schematic of computing distance among kernels of layer $i$. The left side is the set of original kernels, and the right side is the set of flipped kernels. The line represents the operation that calculates the distance of the kernels at both ends of the line.}
		\label{Schematic}
	\end{figure}
	
	To visualize for calculating average kernel similarity ${\bf{S}}\left(  \bullet  \right)$, we can observe Fig. \ref{Schematic}. There are two sets in Fig. \ref{Schematic} distinguished by boxes, which are the original kernel set ${K_i} = \left\{ {{K_{i,1}},{K_{i,2}}, \cdots ,{K_{i,j}}} \right\}$ and the flipped kernel set $K_i^{'} = \left\{ {K_{i,1}^{'},K_{i,2}^{'}, \cdots ,K_{i,j}^{'}} \right\}$ respectively, $i = 1,2, \ldots ,N$ and $j = 1,2 \ldots ,{B_i}$, where ${K_{i,j}}$ represents the $jth$ kernel of the convolution layer $i$ and $K_{i,j}^{'}$ represents the $jth$ flipped kernel of the convolution layer $i$. The dots in Fig. \ref{Schematic} represent kernels and the lines represent the operation that calculates the distance of kernels at both ends. Each convolution layer computes the distance ${B_i} \times {B_i}$ times, and the results are finally averaged to obtain the average kernel similarity of the layer. The formal description of ${\bf{S}}\left(  \bullet  \right)$ is as follows:

	\begin{table*}[hbtp]
		\renewcommand{\arraystretch}{1.3}
		\caption{Architectures of ResNet}
		\centering
		\begin{tabular}{cccccc}
			\hline
			Stage & ResNet-18 & ResNet-34 & ResNet-50 & ResNet-101 & ResNet-152 \\ \hline
			1 & -- & -- & -- & -- & -- \\
			2     & $\left[ \begin{array}{l}
				3 \times 3,64\\
				3 \times 3,64
			\end{array} \right] \times 2$ & $\left[ \begin{array}{l}
				3 \times 3,64\\
				3 \times 3,64
			\end{array} \right] \times 3$ & $\left[ \begin{array}{l}
				1 \times 1,64\\
				3 \times 3,64\\
				1 \times 1,256
			\end{array} \right] \times 3$ & $\left[ \begin{array}{l}
				1 \times 1,64\\
				3 \times 3,64\\
				1 \times 1,256
			\end{array} \right] \times 3$ & $\left[ \begin{array}{l}
				1 \times 1,64\\
				3 \times 3,64\\
				1 \times 1,256
			\end{array} \right] \times 3$\\
			3     & $\left[ \begin{array}{l}
				3 \times 3,128\\
				3 \times 3,128
			\end{array} \right] \times 2$ & $\left[ \begin{array}{l}
				3 \times 3,128\\
				3 \times 3,128
			\end{array} \right] \times 4$ & $\left[ \begin{array}{l}
				1 \times 1,128\\
				3 \times 3,128\\
				1 \times 1,512
			\end{array} \right] \times 4$ & $\left[ \begin{array}{l}
				1 \times 1,128\\
				3 \times 3,128\\
				1 \times 1,512
			\end{array} \right] \times 4$ & $\left[ \begin{array}{l}
				1 \times 1,128\\
				3 \times 3,128\\
				1 \times 1,512
			\end{array} \right] \times 8$\\
			4     & $\left[ \begin{array}{l}
				3 \times 3,256\\
				3 \times 3,256
			\end{array} \right] \times 2$ & $\left[ \begin{array}{l}
				3 \times 3,256\\
				3 \times 3,256
			\end{array} \right] \times 6$ & $\left[ \begin{array}{l}
				1 \times 1,256\\
				3 \times 3,256\\
				1 \times 1,1024
			\end{array} \right] \times 6$ & $\left[ \begin{array}{l}
				1 \times 1,256\\
				3 \times 3,256\\
				1 \times 1,1024
			\end{array} \right] \times 23$ & $\left[ \begin{array}{l}
				1 \times 1,256\\
				3 \times 3,256\\
				1 \times 1,1024
			\end{array} \right] \times 36$\\
			5     & $\left[ \begin{array}{l}
				3 \times 3,512\\
				3 \times 3,512
			\end{array} \right] \times 2$ & $\left[ \begin{array}{l}
				3 \times 3,512\\
				3 \times 3,512
			\end{array} \right] \times 3$ & $\left[ \begin{array}{l}
				1 \times 1,512\\
				3 \times 3,512\\
				1 \times 1,2048
			\end{array} \right] \times 3$ & $\left[ \begin{array}{l}
				1 \times 1,512\\
				3 \times 3,512\\
				1 \times 1,2048
			\end{array} \right] \times 3$ & $\left[ \begin{array}{l}
				1 \times 1,512\\
				3 \times 3,512\\
				1 \times 1,2048
			\end{array} \right] \times 3$\\ \hline
		\end{tabular}
		\label{AR}
	\end{table*}
	
	\begin{equation}
		{{\bf{S}}\left( {{L_i}} \right) = \frac{{\sum\limits_{{j^{'}} = 1}^{{B_i}} {\sum\limits_{j = 1}^{{B_i}} {\left| {cos\left( {{K_{i,j}},K_{i,{j^{'}}}^{'}} \right)} \right|} } }}{{{{\left( {{B_i}} \right)}^2}}}}
		\label{average}
	\end{equation}
	
	\noindent where ${K_{i,j}}$ is the $jth$ kernel of the layer $i$, $K_{i,{j^{'}}}^{'}$ is the flipped kernel of ${K_{i,j}}$, and $cos\left(\bullet\right)$ is cosine distance. The size of these kernels' parameter is $\left\langle {{C_i},{H_i},{W_i}} \right\rangle$. For all $j = 1,2,...,{B_i}$, we calculate the flipped kernel:
	
	\begin{equation}
		{K_{i,j}^{'} = {\bf{T}}\left( {{K_{i,j}}} \right)}
		\label{ktk}
	\end{equation}
	
	\noindent where ${\bf{T}}\left(  \bullet  \right)$ is a flipping operation, which can make the kernel keep the order of channels while flipping the kernel in the horizontal direction in this paper.

	The detailed description of ${\bf{T}}\left( {{K_{i,j}}} \right)$ is as follows: Firstly, divide the kernel ${K_{i,j}}$ into ${C_i}$ two-dimensional matrices according to the channel order. Then, multiply these two-dimensional matrices by the inversion matrix ${T_2} = \left( {\begin{array}{*{20}{c}}
			0&0&1\\
			0&1&0\\
			1&0&0
	\end{array}} \right)$ successively, so that the matrix flips horizontally with the elements in the second column as the axis. For example, when a slice matrix of ${K_{i,j}}$ is $\left( {\begin{array}{*{20}{c}}
			1&2&3\\
			4&5&6\\
			7&8&9
	\end{array}} \right)$, multiply it by ${T_2}$ to get $\left( {\begin{array}{*{20}{c}}
			3&2&1\\
			6&5&4\\
			9&8&7
	\end{array}} \right)$. Finally, splicing these two-dimensional matrices back to become the mirror kernel $K_{i,j}^{'}$ according to the original channel order, which size is $\left\langle {{C_i},{H_i},{W_i}} \right\rangle$.
	
	We use $cos\left( {{K_{i,j}}, K_{i,{j^{'}}}^{'}} \right)$ to calculate the cosine distance of ${K_{i,j}}$ and $K_{i,j}^{'}$, which pulls the kernels into two vectors separately, to calculate cosine distance so that the range of value obtained is $\left[ { - 1,1} \right]$. When the value of cosine distance is $ - 1$, the two kernels are opposite; when the cosine distance value is $0$, the two kernels are orthogonal; when the cosine distance is $1$, the two kernels are precisely the same. Since a pair of opposite kernels is the same as a pair of identical kernels in terms of whether they are chiral or not, we take the absolute value of each distance. The greater the cosine distance between two objects, the lower the similarity. 
	
	\begin{figure}[hbtp]
		\centerline{\includegraphics[width=3in]{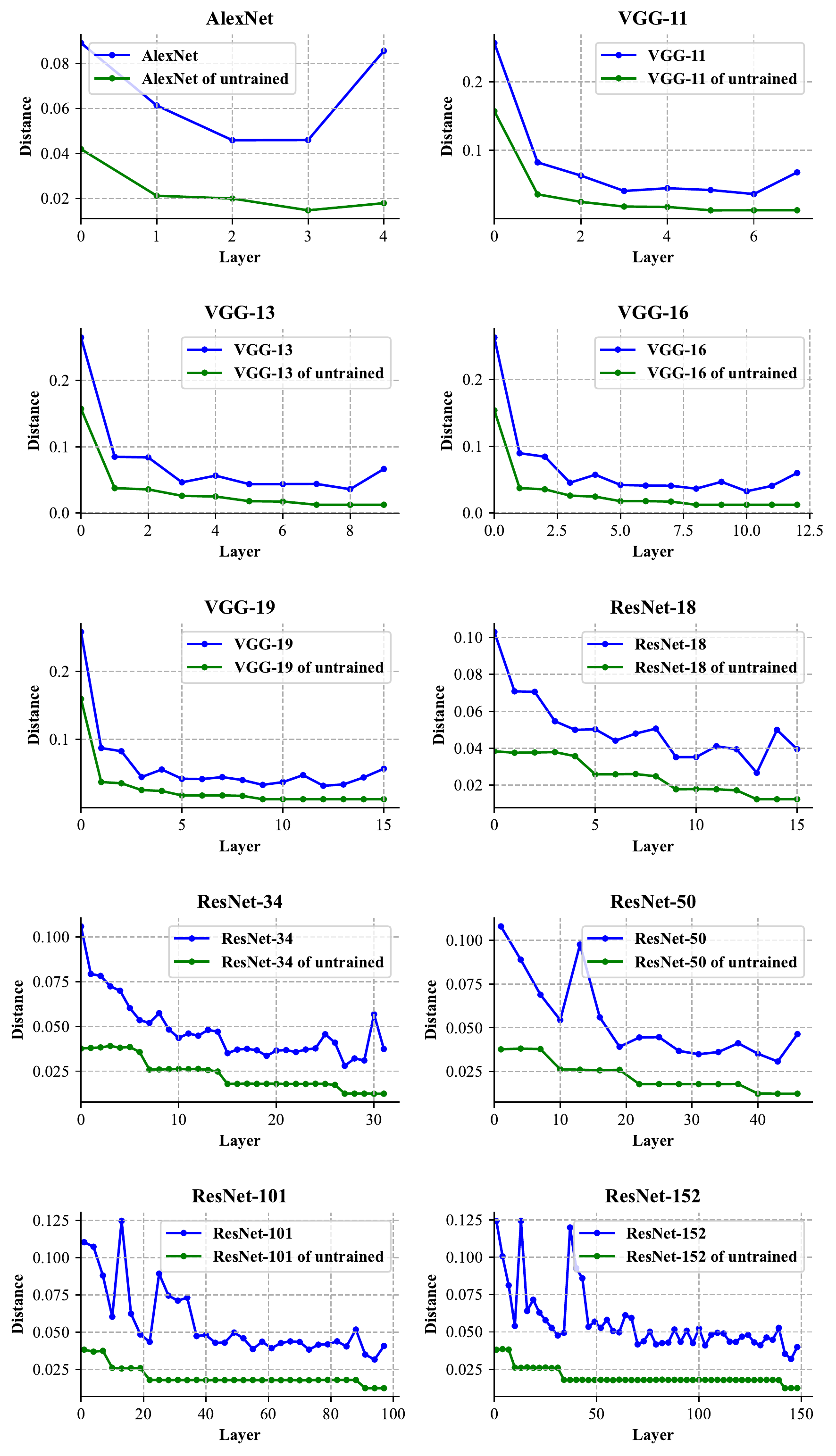}}
		\caption{Influence of models training on average kernel similarity.}
		\label{pre and un}
	\end{figure}
	
	\section{Experiments}
	The pre-trained models used in the experiment of this paper are all from the \textit{models} module of \textit{torchvision} whose version is 0.10.0+cu102. In this section, we use the measure described in Section \ref{third} to test AlexNet \cite{krizhevsky2017imagenet}, VGG \cite{simonyan2014very}, and ResNet \cite{he2016deep}.
	
	\subsection{Implementation Details}
	In this section, the model parameters of \textit{alexnet}, \textit{vgg11}, \textit{vgg13}, \textit{vgg16}, \textit{vgg19}, \textit{resnet18}, \textit{resnet34}, \textit{resnet50}, \textit{resnet101} and \textit{resnet152} in the \textit{models} module were used. As shown in Table \ref{AAA} and Table \ref{AR}, since the ResNet is designed according to stage-block division unlike AlexNet and VGG, the convolution layers of ResNet are in different blocks within the same stage. Furthermore, except for the first convolution layer in each block, the numbers of the input channel and output channel are consistent. In view of stage 1 of ResNet being used for input preprocessing, this stage is excluded from the analysis in this paper. Meanwhile, in this section, we only analyze the kernels with the size that can be flipped, which also makes the number of output channels in each block consistent in the same stage.
	
	\subsection{Analysis of Experiments}
	\textbf{Feature chirality exits.} As shown in Fig. \ref{pre and un}, The distance between the kernels of each model increases significantly after training, which means that the average kernel similarity decreases. According to our definition, feature chirality is ubiquitous in deep learning models.
	
	\begin{figure}[hbtp]
		\centerline{\includegraphics[width=3in]{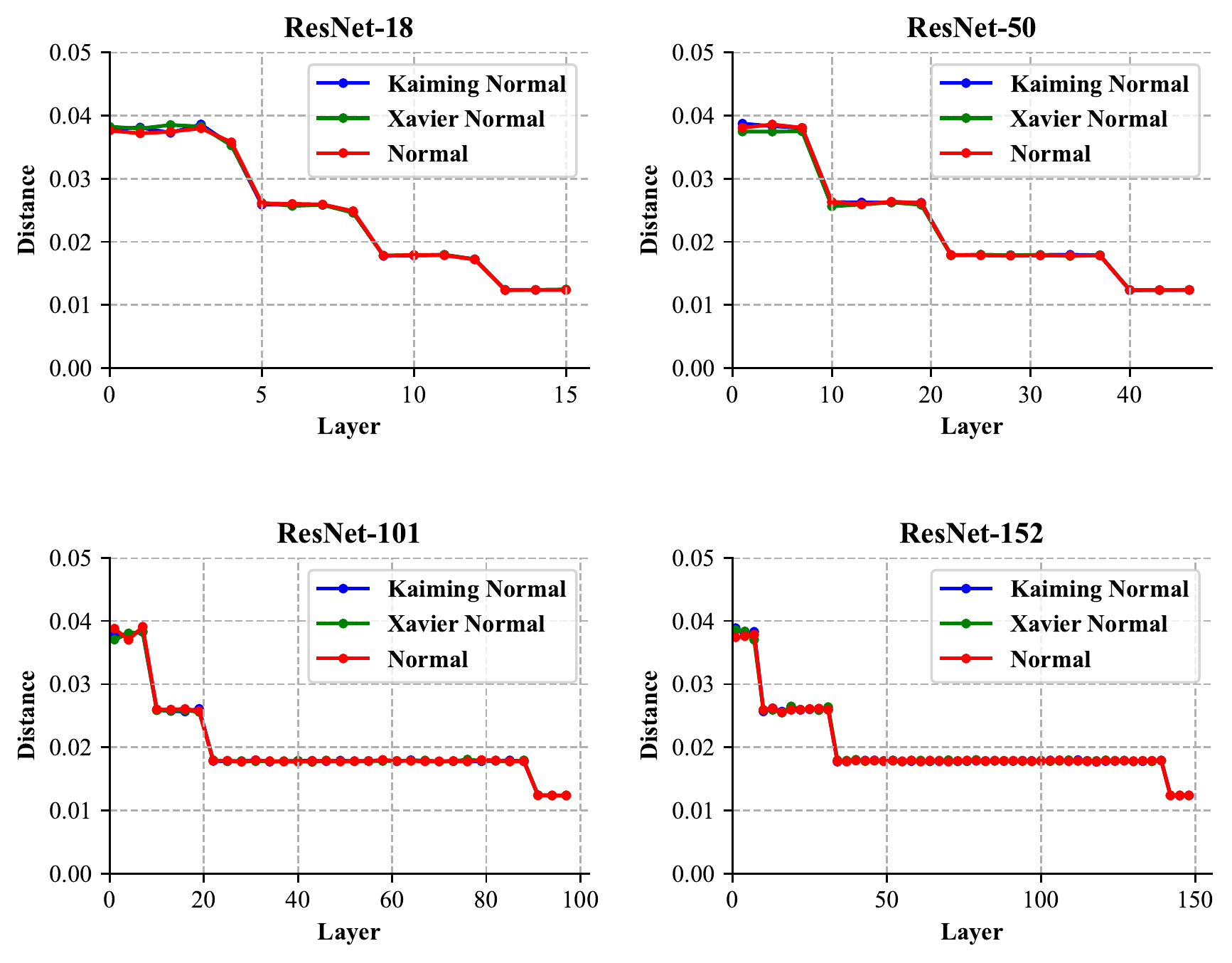}}
		\caption{Influence of initialization methods on average kernel similarity.}
		\label{initialization}
	\end{figure}
	
	\textbf{The initialization method does not affect the initial average kernel similarity of the model.} We conduct experiments on different initialization methods to understand their effects on the initial average kernel similarity of the model. We compare the cases where the initialization method is \textit{kaiming normal}, \textit{xavier normal}, and \textit{normal}. As shown in Fig. \ref{initialization}, the difference in the results is slight. Considering that the initialization methods use random parameters, the initialization methods of the models do not affect the initial average kernel similarity of the model.

	\textbf{The variation trend of average kernel similarity is consistent with that of the serial models.} Since the number of convolution layers between models is different, to facilitate finding the variation trend of average kernel similarity, the results of ResNet and VGG are divided convolution layers into five stages according to the stage design of ResNet, and the results of convolution layers within the same layer are uniformly stretched horizontally. As shown in Fig. \ref{series}, the average kernel similarity of VGG and ResNet increased with the increase of stage, other than the result of AlexNet (as shown in Fig. \ref{pre and un}) increase first and then decrease. Therefore, the distance of kernels of the ResNet decreased with the increase of the stage (which can be observed more clearly in the untrained results, as shown in Fig. \ref{initialization}), VGG’s distance of kernels is similar to the decline of the logarithmic function curve. 
		
	\begin{figure}[hbtp]
		\centerline{\includegraphics[width=3in]{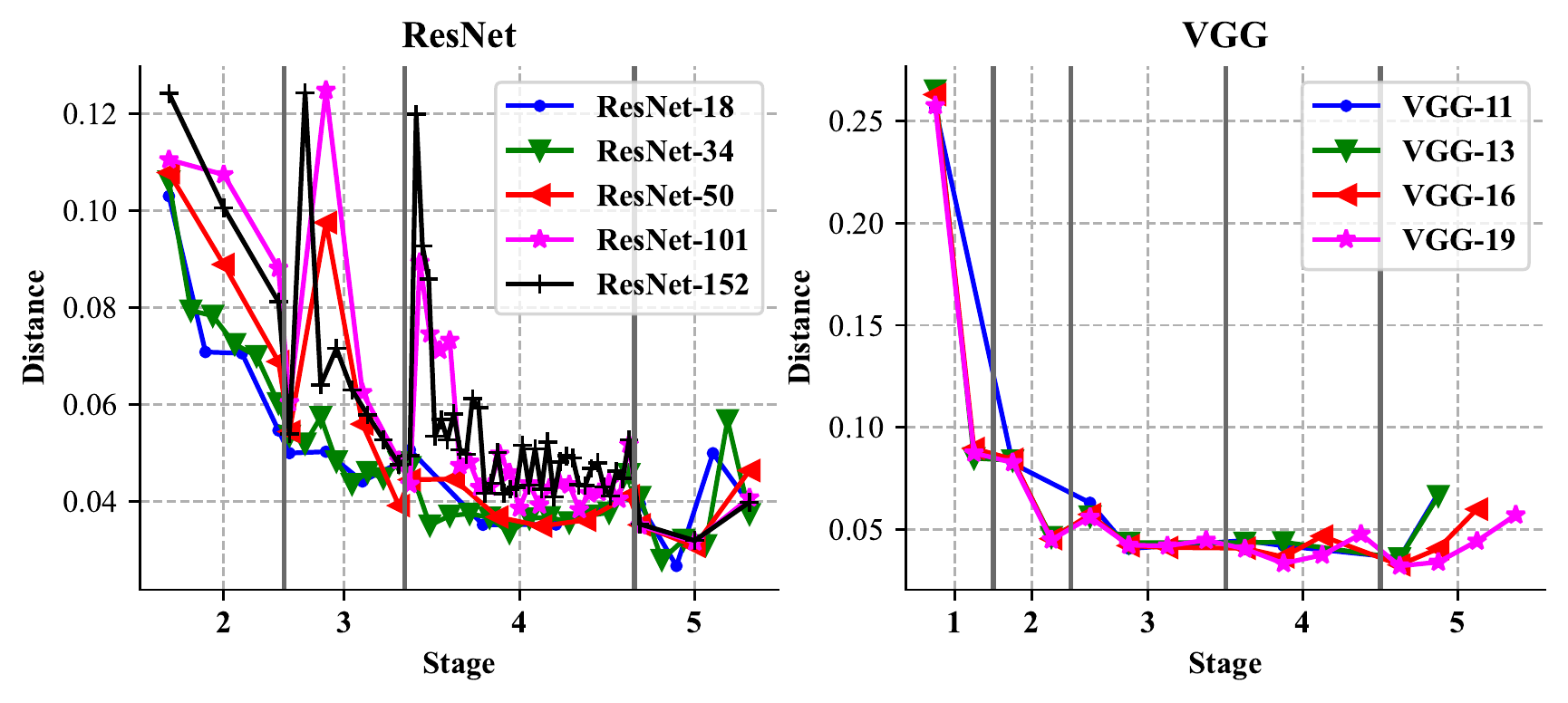}}
		\caption{Trend of average kernel similarity of the models in same series.}
		\label{series}
	\end{figure}
	
	Two reasons for the result in AlexNet can be associated. Firstly, the phenomenon of improving the average kernel similarity in the last few convolution layers is common in the various models. However, it is more obvious in AlexNet because of the lower average kernel similarities and the fewer convolution layers. Secondly, in Fig. \ref{series}, the ranking of average kernel similarities of ResNets is roughly consistent with its model accuracy, while the accuracy of the AlexNet model is not higher than that of the other two series models.

	\subsection{Ablation experiment}
	To understand the effect of the flip or not in the measure of the average kernel similarity of the model, we conducted an ablation experiment. We replace the set of flipped kernels in Eq. (\ref{average}) with the set of original kernels. In other words, we calculate the distance between two sets of original kernels. The obtained results are shown in Fig. \ref{ablation}. Using two sets of original kernels will make the distance of kernels decrease a little, but the trend remains the same.
	
	\begin{figure}[hbtp]
		\centerline{\includegraphics[width=3in]{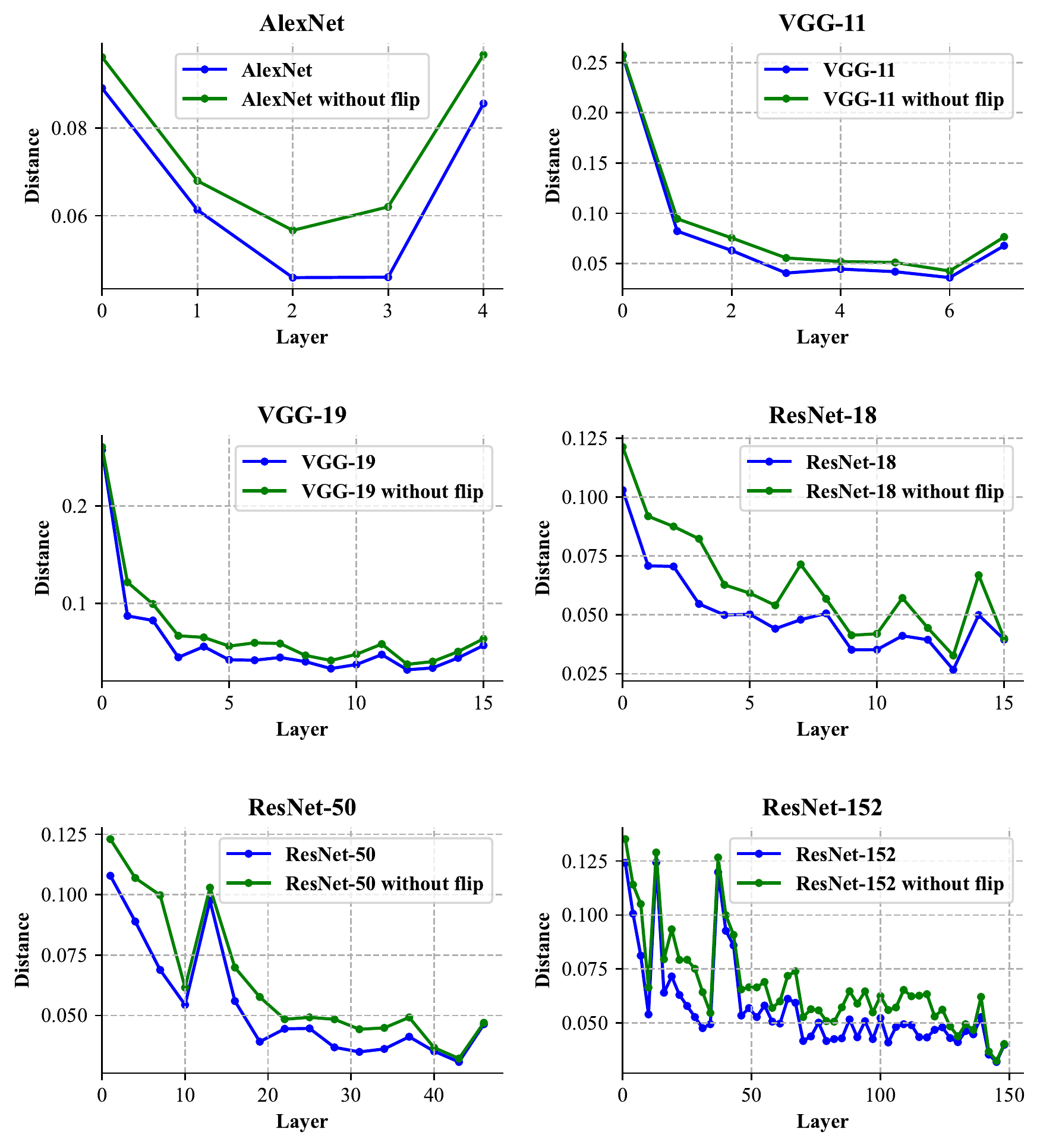}}
		\caption{Influence of not flipping the kernel.}
		\label{ablation}
	\end{figure}

	\section{Conclusion}
	This paper studies the chiral phenomena in deep learning models. We proposed feature chirality and measured it on AlexNet, VGG, and ResNet. We found that feature chirality is ubiquitous in models and presents certain regularity, which will be one of our future studies. Furthermore, the feature chirality indicates the accuracy ranking of the same series of models, which may provide a certain reference for model evaluation in the future. The performance of feature chirality on the same series of models can also help us to distinguish the types of models and judge whether the models are trained or not, which also reflects the unique properties of different series of models from the side. At present, there are few pieces of research on this topic, lacking sufficient consensus and results. The findings of this work may be more inspiring contributions to the future work of other researchers on the interpretability of deep learning models from the perspective of chirality and model parameters optimization.
	
	\balance
	\bibliographystyle{IEEEtran}
	\bibliography{refs}
	
\end{document}